%% file: paper_main.tex
\documentclass[10pt,twocolumn]{article} 
\usepackage[a4paper, left=15mm, top=20mm, bottom=20mm, right=12mm]{geometry}

\input{srctex/head.tex}

\title{\LARGE \bf Towards a Generic Diver-Following Algorithm: Balancing Robustness and Efficiency in Deep Visual Detection}%
\author{
    Md Jahidul Islam$^{1}$, Michael Fulton$^{2}$ and Junaed Sattar$^{3}$%
    \thanks{The authors are with the Interactive Robotics and Vision Laboratory, Department of Computer Science and Engineering, University of Minnesota- Twin Cities, US.  {\tt\footnotesize \{$^{1}$islam034, $^{2}$fulto081, $^{3}$junaed\}@umn.edu}%
    }
}
\date{Preprint (v1) of ICRA 2019 submission}

\begin{document}

\maketitle
\thispagestyle{empty}
\pagestyle{empty}

\input{srctex/abstract.tex}
\input{srctex/intro.tex}
\input{srctex/rel.tex}

\input{srctex/metho.tex}

\input{srctex/res.tex}

\input{srctex/con.tex}

\bibliographystyle{plain}
\bibliography{refs}

\end{document}

%% file: srctex/head.tex
\usepackage{graphicx} 
\usepackage{subcaption}
\usepackage{amsmath} 
\usepackage{amssymb}  
\usepackage{csquotes}

\usepackage{array}
\usepackage{url} 
\usepackage{enumerate}
\usepackage{multirow} 
\usepackage{float}
\usepackage{paralist}
\usepackage{amsfonts}
\usepackage{longtable}

\newcolumntype{L}[1]{>{\raggedright\let\newline\\\arraybackslash\hspace{0pt}}m{#1}}
\newcolumntype{C}[1]{>{\centering\let\newline\\\arraybackslash\hspace{0pt}}m{#1}}
\newcolumntype{R}[1]{>{\raggedleft\let\newline\\\arraybackslash\hspace{0pt}}m{#1}}

\graphicspath{ {images/} }

\newcommand{\nostarnote}[1]{}
\newcommand{\baad}[1]{} 
\usepackage[size=small]{caption}

%% file: srctex/abstract.tex
\begin{abstract}
This paper explores the design and development of a class of robust diver-following algorithms for autonomous underwater robots. 
By considering the operational challenges for underwater visual tracking in diverse real-world settings, we formulate a set of desired features of a generic diver following algorithm. 
We attempt to accommodate these features and maximize general tracking performance by exploiting the state-of-the-art deep object detection models. 
We fine-tune the building blocks of these models with a goal of balancing the trade-off between robustness and efficiency in an on-board setting under real-time constraints. 
Subsequently, we design an architecturally simple Convolutional Neural Network (CNN)-based diver-detection model that is much faster than the state-of-the-art deep models yet provides comparable detection performances. 
In addition, we validate the performance and effectiveness of the proposed diver-following modules through a number of field experiments in closed-water and open-water environments.
\end{abstract}

%% file: srctex/intro.tex
\section{INTRODUCTION}
Underwater applications of autonomous underwater robots range from inspection and surveillance to data collection and mapping tasks. Such missions often require a team of divers and robots to collaborate for successful completion. Without sacrificing the generality of such applications, we can consider a single-robot setting where a human diver leads the task and interacts with the robot which follows the diver at certain stages of the mission. Such situations arise in numerous important applications such as submarine pipeline and ship-wreck inspection, marine life and seabed monitoring, and many other underwater exploration activities~\cite{sattar2008enabling}. Although following the diver is not the primary objective in these applications, it significantly simplifies the operational loop and reduces the associated overhead by eliminating the necessity of tele-operation.   

Robust underwater visual perception is generally challenging due to marine artifacts~\cite{sattar2006performance} such as poor
visibility, variations in illumination, suspended particles, etc. Additionally, color distortion and scarcity of salient visual features make it harder to robustly detect and accurately follow a diver in arbitrary directions. Moreover, divers' appearances to the robot vary greatly based on their swimming styles, choices of wearables, and relative orientations with respect to the robot. These problems are exacerbated underwater since both the robot and diver are suspended in a six-degrees-of-freedom (6DOF) environment. Consequently, classical model-based detection algorithms fail to achieve good generalization performance~\cite{islam2017mixed, fabbri2018enhancing}. On the other hand, model-free algorithms incur significant target drift~\cite{shkurti2017underwater} under such noisy conditions. 

\begin{figure}[h]
\centering
\includegraphics [width=\linewidth]{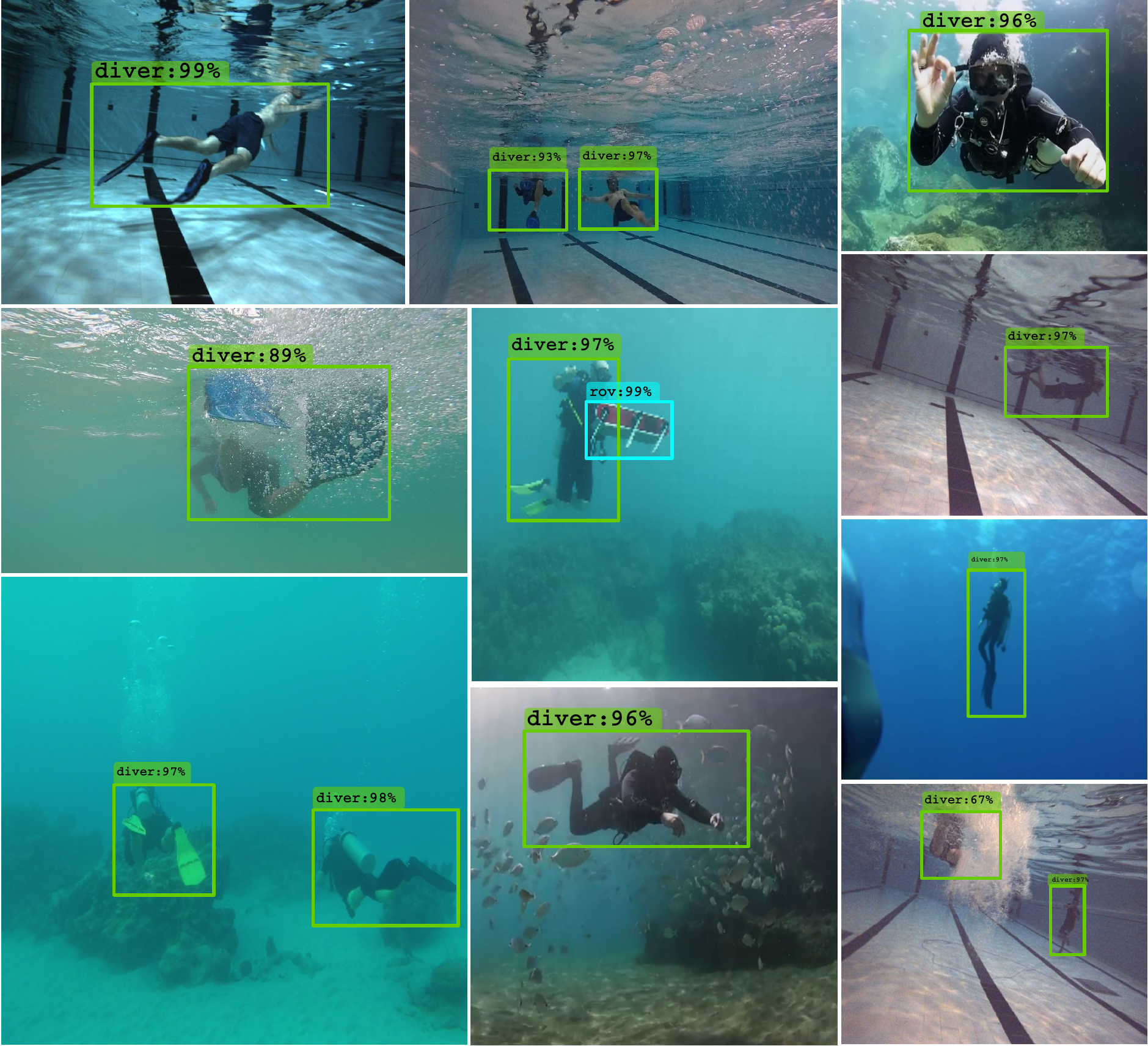} 
\caption{Snapshots of a set of diverse first-person views of the robot from different diver-following scenarios. Notice the variation in appearances of the divers and possible noise or disturbances in the scene over different scenarios. The rectangles and text overlaid on the figures are the outputs generated by our model at test time.}%
\label{fig:m1}
\end{figure}%

In this paper, we address the inherent difficulties of underwater visual detection by designing a class of diver-following algorithms that are:  
\begin{inparaenum}[$a)$]
\item invariant to color (of divers' body/wearables \cite{sattar2007your}),
\item invariant to divers' relative motion and orientation,
\item robust to noise and image distortions~\cite{fabbri2018enhancing}, and 
\item reasonably efficient for real-time deployment.
\end{inparaenum}
We exploit the current state-of-the-art object detectors to accommodate these features and maximize the generalization performance for diver detection using RGB images as input. Specifically, we use the following four models: Faster R-CNN~\cite{renNIPS15fasterrcnn} with Inception V2 \cite{szegedy2016rethinking} as a feature extractor, Single Shot MultiBox Detector (SSD)~\cite{liu2016ssd} with MobileNet V2 \cite{sandler2018inverted, howard2017mobilenets} as a feature extractor, You Only Look Once (YOLO) V2~\cite{redmon2016yolo9000}, and Tiny YOLO \cite{tinyYOLO}. These are the fastest (in terms of processing time of a single frame) among the family of current state-of-the-art models~\cite{tfzoo} for general object detection. We train these models using a rigorously prepared dataset containing sufficient training instances to capture the variabilities of underwater visual sensing. 

Subsequently, we design an architecturally simple (\textit{i.e.}, sparse) deep model that is computationally much faster than the state-of-the-art diver detection models. The faster running time ensures real-time tracking performance with limited on-board computational resources. We also demonstrate its effectiveness in terms of detection performances compared to the state-of-the-art models through extensive quantitative experiments. 
We then validate these results with a series of field experiments. Based on our design, implementation, and experimental findings, we make the following contributions in this paper:
\begin{itemize}
\item We attempt to overcome the limitations of existing model-based diver-following algorithms using state-of-the-art deep object detection models. These models are trained on comprehensive datasets to deal with the challenges involved in underwater visual perception\footnote{The dataset and trained models will be made available for academic research purposes}. 

\item In addition, we design a CNN-based diver detection model to balance the trade-offs between robustness and efficiency. 
The proposed model provides considerably faster running time, in addition to achieving detection performances comparable to the state-of-the-art models.

\item Finally, we validate the effectiveness of the proposed diver-following methodologies through extensive experimental evaluations. A number of field experiments are performed both in open-water and closed-water (\textit{i.e.}, oceans and pools, respectively) environments in order to demonstrate their real-time tracking performances. 
\vspace{1mm}
\end{itemize}

Furthermore, we demonstrate that the proposed models can be extended for a wide range of other applications such as human-robot communication~\cite{islam2018dynamic}, robot convoying~\cite{shkurti2017underwater}, cooperative localization~\cite{bahr2009cooperative,rekleitis2003probabilistic}, etc. The state-of-the-art detection performance, fast running time, and architectural portability are the key features of these models, which make them suitable for 
underwater human-robot collaborative applications.

%% file: srctex/rel.tex
\section{RELATED WORK}
\begin{figure}
 \centering
    \includegraphics[width=0.9\linewidth]{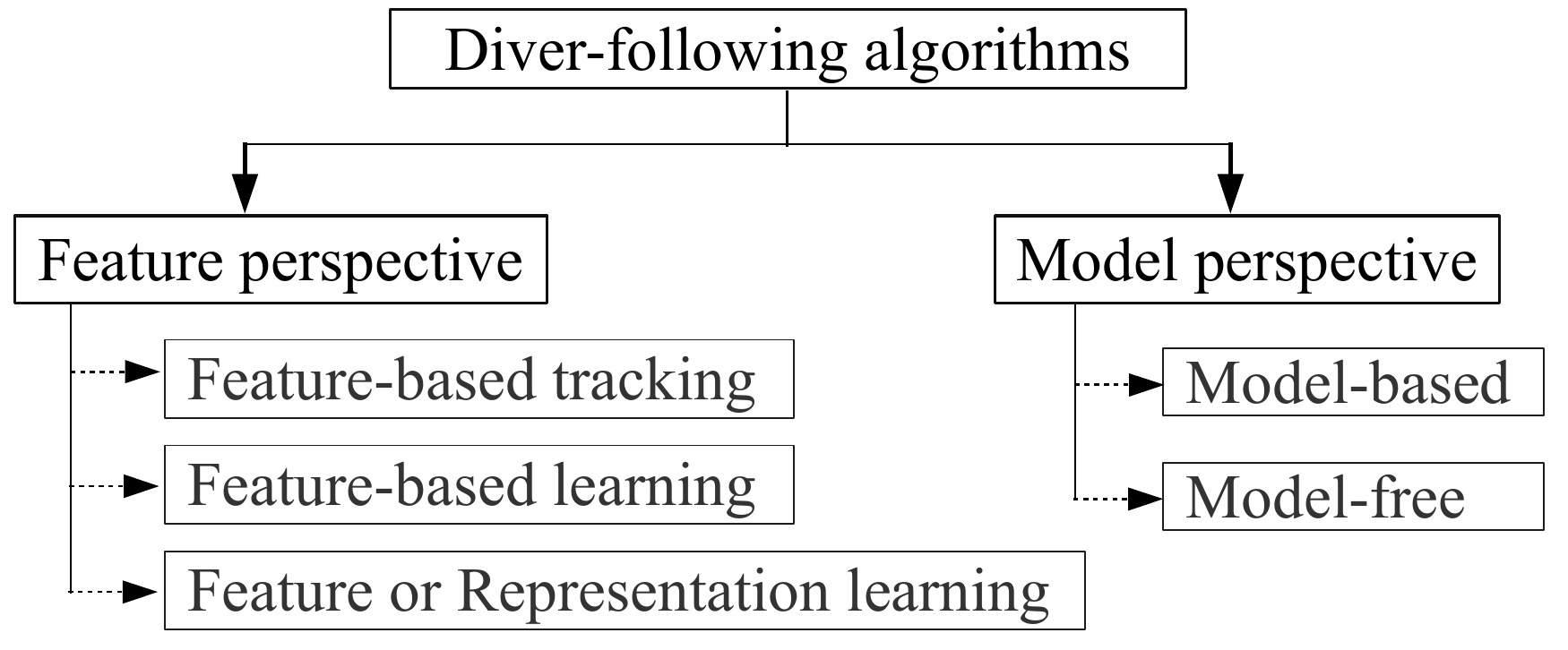}
 \caption{An algorithmic categorization of the visual perception techniques used for diver-following~\cite{islam2018person}}
 \label{fig:perc}
\end{figure}

A categorization of the vision-based diver-following algorithms is illustrated in Figure \ref{fig:perc}. Based on algorithmic usage of the input features, they can be grouped as feature-based tracking, feature-based learning, and feature or representation learning algorithms. On the other hand, they can be categorized into model-based and model-free techniques based on whether or not any prior knowledge about the appearance or motion of the diver is used for tracking. 

\subsection{Model Perspective}
In model-free algorithms, no prior information about the target (\textit{e.g.}, diver's motion model, color of wearables, etc.) is used for tracking. These algorithms are initialized arbitrarily and then iteratively learn to track the target in a semi-supervised fashion~\cite{yu2008online}. TLD (``tracking-learning-detection'')  trackers~\cite{kalal2012tracking} and optical flow-based trackers~\cite{shin2005optical} are the most commonly used model-free algorithms for general object tracking. The TLD trackers train a detector using positive and negative feedback that are obtained from image-based features. In contrast, the optical flow-based methods estimate the motion of each pixel by solving the Horn and Schunck formulation~\cite{inoue1992robot}. Although model-free techniques work reasonably well in practice for general object tracking, they often suffer from tracking drift caused by the accumulation of detection errors over time.

On the other hand, model-based algorithms use prior knowledge about the divers' motion and appearances in order to formulate a model in the input feature-space. Iterative search methods are then applied to find the target model in the feature-space~\cite{islam2017mixed}. Machine learning techniques are also widely used to learn the diver-specific features~\cite{islam2018person,sattar2009robust} and predict the target location in the feature-space. Performance of the model-free algorithms depend on comprehensiveness of the model descriptors and the underlying input feature-space. Hence, they require careful design and thorough training processes to ensure good tracking performance.

\subsection{Feature Perspective}
Simple feature-based trackers~\cite{sattar2006performance,sattar2005visual} are often practical choices for autonomous diver-following due to their operational simplicity and computational efficiency. For instance, color-based trackers perform binary image thresholding based on the color of a diver's flippers or suit. The thresholded binary image is then refined to track the centroid of the target (diver) using algorithms such as mean-shift, particle filters, etc. Optical flow-based methods can also be utilized to track divers' motion in the spatio-temporal volume~\cite{islam2018person,inoue1992robot}. 

Since color distortions and low visibility issues are common in underwater settings, frequency-domain signatures of divers' swimming patterns are often used for reliable detection. Specifically, intensity variations in the spatio-temporal volume caused by a diver's swimming gait generate identifiable high-energy responses in the $1$-$2$Hz frequency range, which can be used for diver detection~\cite{sattar2009underwater}. Moreover, the frequency-domain signatures can be combined with the spatial-domain features for robust diver tracking. For instance, in~\cite{islam2017mixed}, a Hidden Markov Model (HMM) is used to track divers' potential swimming trajectories in the spatio-temporal domain, and then frequency-domain features are utilized to detect the diver along those trajectories.

Another class of approaches use machine learning techniques to approximate the underlying function that relates the input feature-space to the target model of the diver. For instance, Support Vector Machines (SVMs) are trained using Histogram of Oriented Gradients (HOG) features \cite{dalal2005histograms} for robust person detection in general. Ensemble methods such as Adaptive Boosting (AdaBoost)~\cite{sattar2009robust} are also widely used as they are computationally inexpensive yet highly accurate in practice. AdaBoost learns a strong tracker from a large number of simple feature-based diver trackers. Several other machine learning techniques have been investigated for diver tracking and underwater object tracking in general~\cite{islam2018person}. One major challenge involved in using these models is to design a set of robust features that are invariant to noise, lighting condition, and other variabilities such as divers' swimming motion and wearables. 

Convolutional Neural Network(CNN)-based deep models improve generalization performance by learning a feature representation from the image-space. The extracted features are used as inputs to the detector (\textit{i.e.}, fully-connected layers); this end-to-end training process significantly improves the detection performance compared to using hand-crafted features. Once trained with sufficient data, these models are quite robust to occlusion, noise, and color distortions~\cite{shkurti2017underwater}. Despite the robust performance, the applicability of these models to real-time applications is often limited due to their slow running time on embedded devices. In this paper, we investigate the performances and feasibilities of the state-of-the-art deep object detectors for diver-following applications. 
We also design a CNN-based model that achieves robust detection performance in addition to ensuring that the real-time operating constraints are met.

%% file: srctex/metho.tex
\section{NETWORK ARCHITECTURE AND DESIGN}

\subsection{State-of-the-art Object Detectors}
\begin{figure*}[!htb]
    \centering
    \begin{subfigure}[t]{\textwidth}
       \centering
        \includegraphics[width=0.9\linewidth]{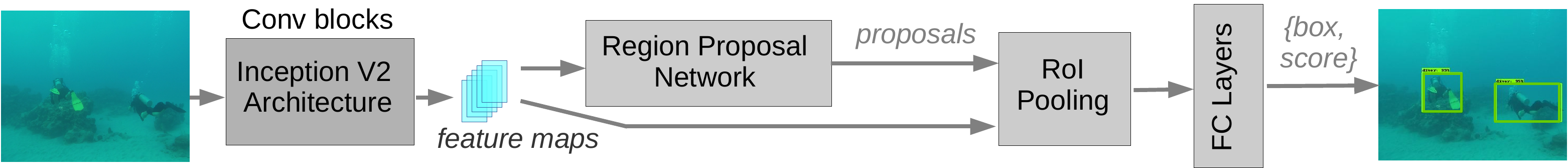}
        \caption{Faster R-CNN with Inception V2}
        \end{subfigure}
        
        \vspace{3mm}
        \begin{subfigure}[t]{\textwidth}
        \centering
        \includegraphics[width=0.9\linewidth]{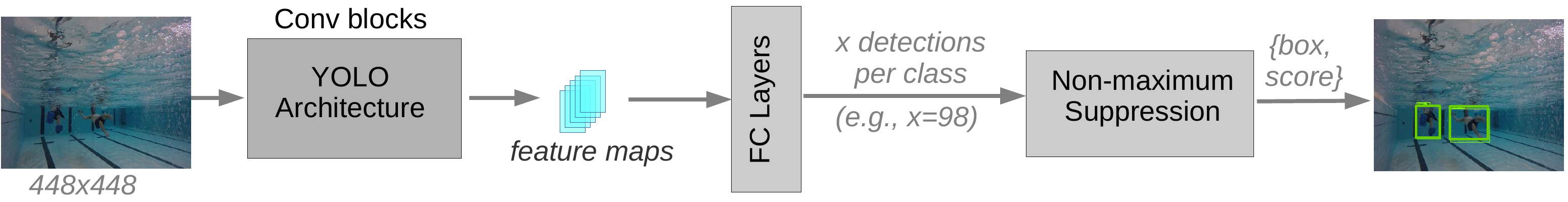}
        \caption{YOLO V2 and Tiny TOLO}
        \end{subfigure}
      
        \vspace{3mm}
        \begin{subfigure}[t]{\textwidth}
        \centering
        \includegraphics[width=0.9\linewidth]{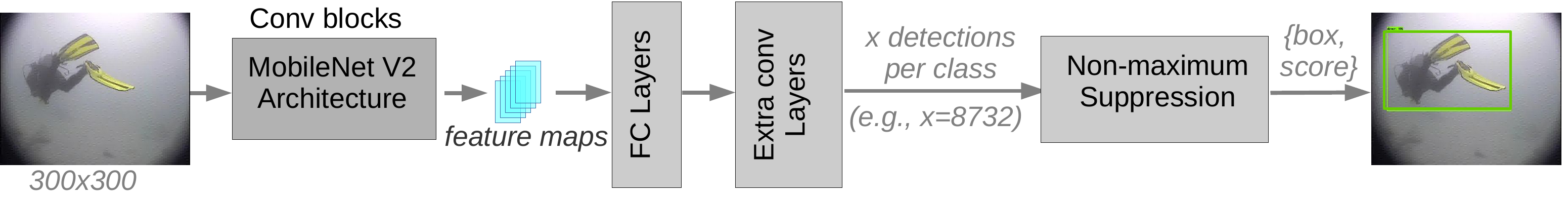}
        \caption{SSD with MobileNet V2}
        \end{subfigure}
        
    \caption{Schematic diagrams of the deep visual models used for diver detection}
    \label{fig:deepGest}
\end{figure*}

We use a Faster R-CNN model, two YOLO models, and an SSD model for diver detection. These are end-to-end trainable models and provide state-of-the-art performances on standard object detection datasets; we refer to~\cite{tinyYOLO,tfzoo} for detailed comparisons of their detection performances and running times. As outlined in Figure \ref{fig:deepGest}, we now briefly discuss their methodologies and the related design choices in terms of major computational components.

\subsubsection{Faster R-CNN with Inception V2}
Faster R-CNN~\cite{renNIPS15fasterrcnn} is an improvement of R-CNN~\cite{Girshick2014RCNN_CVPR} that introduces a Region Proposal Network (RPN) to make the whole object detection network end-to-end trainable. The RPN uses the last convolutional feature-maps to produce region proposals which are then fed to the fully connected layers for the final detection. The original implementation of Faster R-CNN uses VGG-16 \cite{simonyan2014very} model for feature extraction. However, we use Inception V2 \cite{szegedy2016rethinking} model for feature extraction instead, as it is known to provide better object detection performances on standard datasets \cite{tfzoo}.

\subsubsection{YOLO V2 and Tiny YOLO}
YOLO models~\cite{redmon2016you,redmon2016yolo9000} formulate object detection as a regression problem in order to avoid using computationally expensive RPNs. They divide the image-space into rectangular grids and predict a fixed number of bounding boxes, their corresponding confidence scores, and class probabilities. Although there are restrictions on the maximum number of object categories, they perform faster than the standard RPN-based object detectors. Tiny YOLO~\cite{tinyYOLO} is a scaled down version of the original model having sparser layers that runs much faster compared to the original model; however, it sacrifices detection accuracy in the process. 

\subsubsection{SSD with MobileNet V2}
SSD (Single-Shot Detector)~\cite{liu2016ssd} also performs object localization and classification in a single pass of the network using the regression trick as in the YOLO~\cite{redmon2016you} model. The architectural difference of SSD with YOLO is that it introduces additional convolutional layers to the end of a base network, which results in improved performances. In our implementation, we use MobileNet V2 \cite{sandler2018inverted} as the base network to ensure faster running time. 

\begin{figure}[h]
\begin{center}
\includegraphics [width=\linewidth]{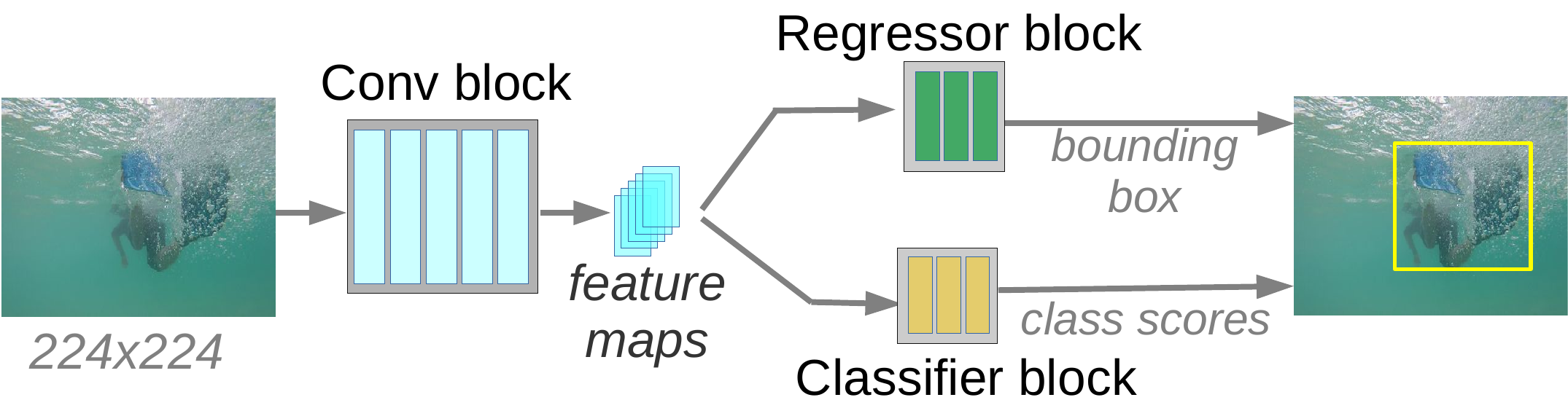}
\caption{A schematic diagram of the proposed CNN-based model for detecting a single diver in the image-space.}
\label{dr_detect}
\end{center}
\end{figure} 

\begin{figure}[h]
\begin{center}
\includegraphics [width=\linewidth]{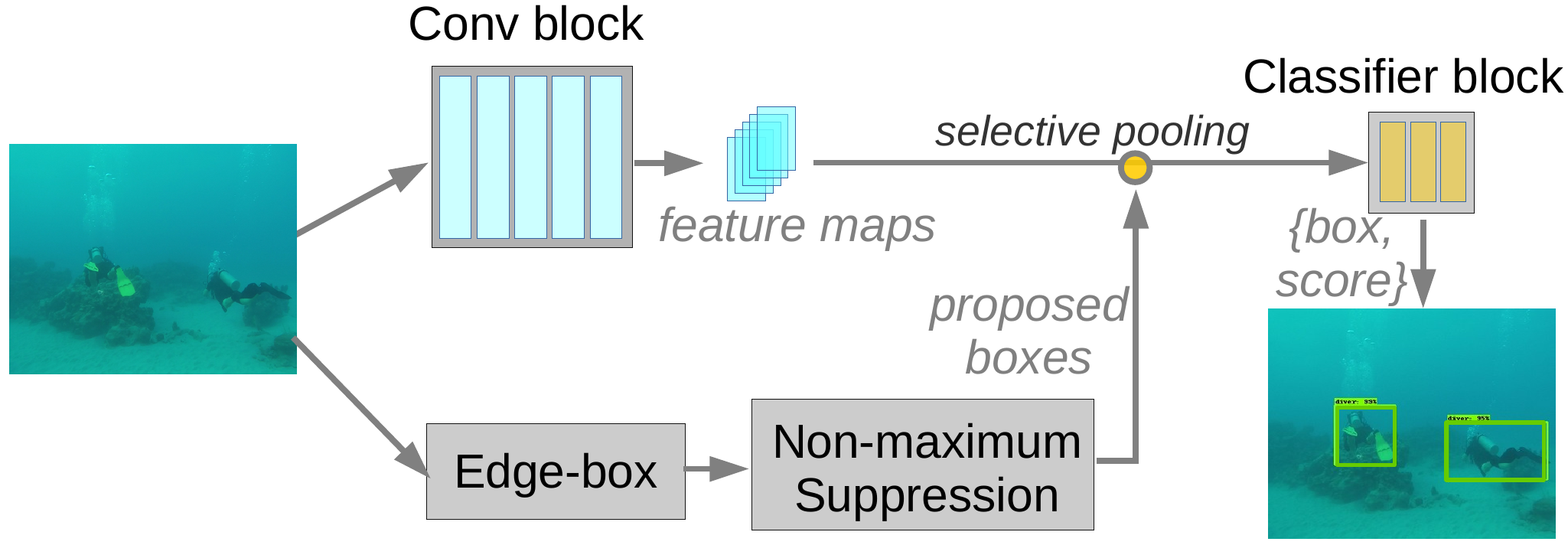}
\caption{Allowing detections of multiple divers in the proposed model using a region selector named Edge-box~\cite{zitnick2014edge}.}
\label{dr_detect_mult}
\end{center}
\end{figure}

\subsection{Proposed CNN-based Model}
Figure \ref{dr_detect} shows a schematic diagram of the proposed CNN-based diver detection model. It consists of three major parts: a convolutional block, a regressor block, and a classifier block. The convolutional block consists of five layers, whereas the classifier and regressor block each consist of three fully connected layers. Detailed network parameters and dimensions are specified in Table \ref{tab:conv}. 

\subsubsection{Design Intuition}
The state-of-the-art deep visual models are designed for  general applications and are trained on standard datasets having a large number of object categories. However, for most underwater human-robot collaborative applications including diver-following, only a few object categories (\textit{e.g.}, diver, robot, coral reefs, etc.) are relevant. We try to take advantage of this by designing an architecturally simpler model that ensures much faster running time in an embedded platform in addition to providing robust detection performance. The underlying design intuitions can be summarized as follows:    

\begin{itemize}
\item The proposed model demonstrated in Figure \ref{dr_detect} is particularly designed for detecting a single diver. Five convolutional layers are used to extract the spatial features in the RGB image-space by learning a set of convolutional kernels. 
\item The extracted features are then fed to the classifier and regressor block for detecting a diver and localizing the corresponding bounding box, respectively. Both the classifier and regressor block consist of three fully connected layers.
\item Therefore, the task of the regressor block is to locate a potential diver in the image-space, whereas the classifier block provides the confidence scores associated with that detection.
\end{itemize}

The proposed model has a sparse convolutional block and uses a three layer regressor block instead of using an RPN. As demonstrated in Table \ref{tab:conv}, it has significantly fewer network parameters compared to the state-of-the-art object detection models.

\begin{table}[h]
\centering
\caption{Parameters and dimensions of the CNN model outlined in Figure \ref{dr_detect}. (convolutional block: conv1-conv5, classifier block: fc1-fc3, regression block: rc1-rc3; n: the number of object categories; *an additional pooling layer was used before passing the conv5 features-maps to fc1)}
\footnotesize
\begin{tabular}{|m{1cm} m{1.5cm} m{1.5cm} cm{1.5cm}|}
\hline
Layer & Input feature-map & Kernel size & Strides  & Output feature-map \\ \hline 
conv1 & 224x224x3 & 11x11x3x64 & [1,4,4,1]   & 56x56x64 \\
pool1 & 56x56x64 & 1x3x3x1 & [1,2,2,1]  & 27x27x64  \\  \hline
conv2 & 27x27x64 & 5x5x64x192 & [1,1,1,1]   & 27x27x192 \\
pool2 & 27x27x192 & 1x3x3x1 & [1,2,2,1]  &  13x13x192  \\ \hline 
conv3 & 13x13x192 & 3x3x192x192 & [1,1,1,1]   & 13x13x192  \\  
conv4 & 13x13x192 & 3x3x192x192 & [1,1,1,1]   & 13x13x192  \\  
conv5 & 13x13x192 & 3x3x192x128 & [1,1,1,1]   & 13x13x128  \\ \hline \hline 
fc1 & 4608x1$^*$ & $-$ & $-$   & 1024x1  \\
fc2 & 1024x1 & $-$ &  $-$  & 128x1  \\
fc3 & 128x1 & $-$ &  $-$  & n  \\ \hline \hline
rc1 & 21632x1 & $-$ & $-$   & 4096x1  \\
rc2 & 4096x1 & $-$ & $-$   & 192x1  \\
rc3 & 192x1 & $-$ & $-$ & 4n  \\ \hline 
\end{tabular}
\label{tab:conv}
\end{table}

\begin{figure*}
\centering
\includegraphics [width=\linewidth]{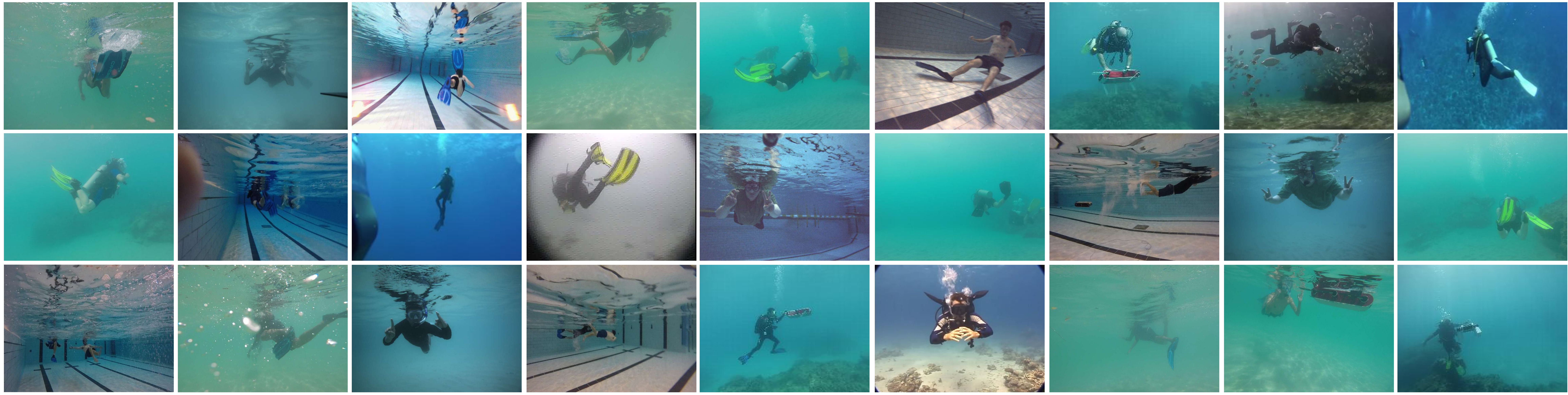} 
\vspace{-3mm}
\caption{A few samples from the training dataset are shown. The annotated training images have class labels (\textit{e.g.}, diver, robot) and corresponding bounding boxes. A total of $30$K of these annotated images are used for supervised training.}
\label{fig:data}
\end{figure*}

\subsubsection{Allowing Multiple Detections}
Although following a single diver is the most common diver-following scenario, detecting multiple divers and other objects is necessary for many human-robot collaborative applications. As shown in Figure \ref{dr_detect_mult}, we add muti-object detection capabilities in our proposed model by replacing the regressor with a region selector. We use the state-of-the-art class-agnostic region selector named Edge-box \cite{zitnick2014edge}. Edge-box utilizes the image-level statistics like edges and contours in order to measure \textit{objectness scores} in various prospective regions in the image-space. 

We use the same convolutional block to extract feature maps. The bounding boxes generated by Edge-box are filtered based on their objectness scores and then non-maxima suppression techniques are applied to get the dominant regions of interest in the image-space. The corresponding feature maps are then fed to the classifier block to predict the object categories. Although we need additional computation for Edge-box, it runs independently and in parallel with the convolutional block; the overall pipeline is still faster than if we were to use an RPN-based object detector model.

%% file: srctex/res.tex
\section{EXPERIMENTS}
We now discuss the implementation details of the proposed networks and present the experimental results.

\subsection{Dataset Preparation}
We performed numerous diver-following experiments in pools and oceans in order to prepare training datasets for the deep models. In addition, we collected data from underwater field trials that are performed by different research groups over the years in pools, lakes, and oceans. This variety of experimental setups is crucial to ensure comprehensiveness of the datasets so that the supervised models can learn the inherent diversity of various application scenarios. We made sure that the datasets contain training examples to capture the following variabilities: 
\begin{itemize}
\item Natural variabilities: changes in visibilities for different sources of water, lighting conditions at varied depths, chromatic distortions, etc.
\item Artificial variabilities: data collected using different robots and cameras.  
\item Human variabilities: different persons and appearances, choice and variations of wearables such as suits, flippers, goggles, etc. 
\end{itemize}

We extracted the robot's camera-feed during these experiments and prepared image-based datasets for supervised training. The images are annotated using the `label-image' software (\url{github.com/tzutalin/labelImg}) by a number of human participants (acknowledged later in the paper) over the period of six months. Few sample images from the dataset are shown in Figure \ref{fig:data}; it contains a total of $30$K images, which are annotated to have class-labels and bounding boxes.

\subsection{Supervised Training Processes}
We train all the supervised deep models on a Linux machine with four GPU cards (NVIDIA\texttrademark{} GTX 1080). TensorFlow~\cite{abadi2016tensorflow} and Darknet~\cite{tinyYOLO} libraries are used for implementation. Once the training is done, the trained inference model (and parameters) is saved and transferred to the robot CPU for validation and real-time experiments.

\begin{figure}[h]
\centering
\includegraphics [width=\linewidth]{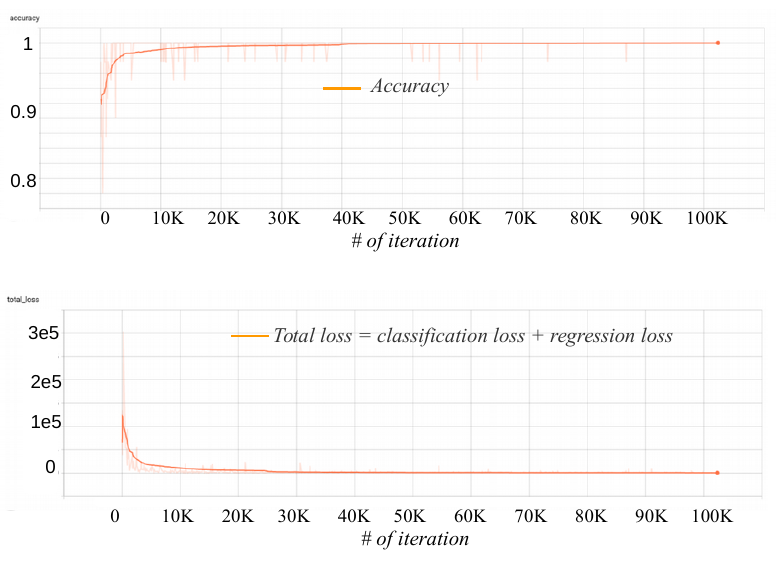} 
\caption{Convergence behavior of the proposed CNN-based model in terms of training accuracy (top) and training loss (bottom).}
\label{fig:train}
\end{figure}

For the state-of-the-art models (Figure \ref{fig:deepGest}), we utilized the pre-trained models for Faster R-CNN, YOLO, and SSD. These models are trained with the recommended configurations provided with their APIs; we refer to~\cite{tinyYOLO,tfzoo} for the detailed processes. On the other hand, our proposed CNN-based models are trained from scratch. Non-supervised pre-training and drop-outs are not used while training. RMSProp \cite{tieleman2012lecture} is used as the optimization function with an initial learning rate of $0.001$. In addition, standard cross-entropy and $L_2$ loss functions are used by the classifier and regressor, respectively. Visualization for the overall convergence behavior of the model is provided in Figure \ref{fig:train}.

\subsection{Performance Evaluation}
We evaluate and compare detection performances of all the models based on standard performance metrics. The test dataset contain $2.2$K images that are chosen from separate field experiments (\textit{i.e.}, they are excluded from the training dataset).

\vspace{1mm}
\subsubsection{Metrics}
We use the following two standard performance metrics: 
\begin{itemize}
\item mAP (mean Average Precision): it is the average of the maximum precisions at different recall values. The precision and recall are defined as $precision = \frac{TP}{TP+FP}$ and $recall = \frac{TP}{TP+FN}$; here, the terms TP, FP, and FN are short forms of True Positive, False Positive, and False Negative, respectively. 

\item IoU (Intersection over Union): it is a measure of how well a model predicts the locations of the objects. It is calculated using the area of overlapping regions of the predicted and ground truth bounding boxes, defined as $IoU = \frac{Area\text{ }of\text{ }overlap}{Area\text{ }of\text{ }union}$ 
\end{itemize} 

As their definitions suggest, mAP measures the detection accuracy, and IoU measures the object localization performance. We also evaluate and compare the running times of the models based on FPS (Frames Per Second), the (average) number of image-frames that a model can process per second. We measure the running times on three different devices: 
\begin{itemize}
\item NVIDIA\texttrademark{} GTX 1080 GPU
\item Embedded GPU (NVIDIA\texttrademark{} Jetson TX2)
\item Robot CPU (Intel\texttrademark{} i3-6100U)
\end{itemize}

\vspace{1mm}
\subsubsection{Results}\label{val_res}
The performances of the diver detection models based on mAP, IoU, and FPS are illustrated in Table \ref{res_com}. The Faster R-CNN (Inception V2) model achieves much better detection performances compared to the other models although it is the slowest  in terms of running time. On the other hand, YOLO V2, SSD (MobileNet V2), and the proposed CNN-based model provide comparable detection performances. Although Tiny YOLO provides fast running time, its detection performance is not as good as the other models. The results demonstrate that the proposed CNN-based model balances the trade-off between detection performances and running time. In addition to the good detection performances, a running time of $6.85$ FPS on the robot CPU and $17.35$ FPS on the embedded GPU validate its applicability in real-time diver-following applications.    

\begin{table}
\footnotesize
\caption{Performance comparison for the diver detection models based on standard metrics.}
\centering
  \begin{tabular}{|m{2.2cm}|c|c|m{0.8cm}|m{0.8cm}|m{0.8cm}|}
  \hline
    \multirow{2}{*}{Models} &
      \multirow{2}{*}{mAP} & \multirow{2}{*}{IoU} &
      \multicolumn{3}{c|}{FPS} \\
      \cline{4-6}
      & (\%) & (\%) & GTX 1080 & Jetson TX2 & Robot CPU \\  
       \hline \hline
       Faster R-CNN $\quad$ (Inception V2) & $71.1$ & $78.3$ & $17.3$ & $2.1$ & $0.52$ \\ \hline
       YOLO V2 & $57.84$ & $62.42$ & $73.3$ & $6.2$ & $0.11$ \\ \hline
       Tiny YOLO & $52.33$ & $59.94$ & $220$ & $20$ & $5.5$ \\ \hline
       SSD $\quad$ (MobileNet V2) & $61.25$ & $69.8$ & $92$ & $9.85$ & $3.8$ \\ \hline
       Proposed CNN-based Model & $53.75$ & $67.4$ & $263.5$ & $17.35$ & $6.85$ \\ \hline
  \end{tabular}
\label{res_com} 
\vspace{-1mm}
\end{table} 

\subsection{Field Experiments}
\subsubsection{Setup}
We have performed several real-world experiments both in closed-water and in open-water conditions (\textit{i.e.}, in pools and in oceans). An autonomous underwater robot of the Aqua~\cite{dudek2007aqua} family is used for testing the diver-following modules. During the experiments, a diver swims in front of the robot in arbitrary directions. The task of the robot is to visually detect the diver using its camera feed and follow behind him/her with a smooth motion.

\subsubsection{Visual Servoing Controller}\label{vizSer}
\begin{figure}[!h]
 \centering
    \includegraphics[width=0.7\linewidth]{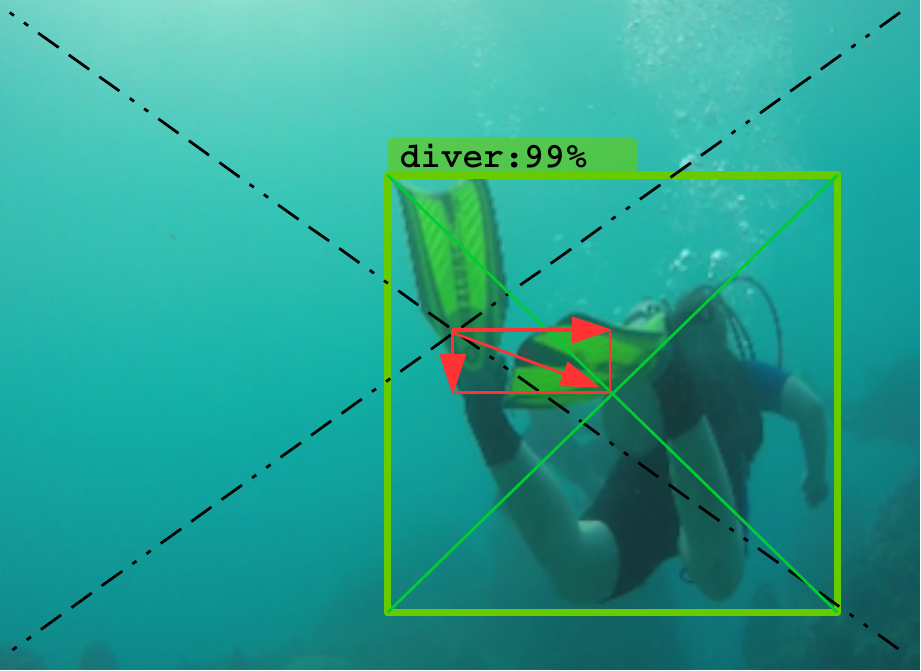}
 \caption{Illustration of how the yaw and pitch commands are generated based on the horizontal and vertical displacements of the center of the detected bounding box}
 \label{fig:serv}
\end{figure}

The Aqua robots have five degrees-of-freedom of control, \textit{i.e.}, three angular (yaw, pitch, and roll) and two linear (forward and vertical speed) controls. In our experiments for autonomous diver-following, we adopt a tracking-by-detection method where the visual servoing~\cite{espiau1992new} controller uses the uncalibrated camera feeds for navigation. The controller regulates the motion of the robot in order to bring the observed bounding box of the target diver to the center of the camera image. The distance of the diver is approximated by the size of the bounding box and forward velocity rates are generated accordingly. Additionally, the yaw and pitch commands are normalized based on the horizontal and vertical displacements of the observed bounding box-center from the image-center (see Figure~\ref{fig:serv}); these navigation commands are then regulated by separate PID controllers. On the other hand, the roll stabilization and hovering are handled by the robot's autopilot module~\cite{meger20143d}. 

\begin{figure*}[h]
    \centering
    \begin{subfigure}[t]{0.35\textwidth}
       \centering
        \includegraphics[width=\linewidth]{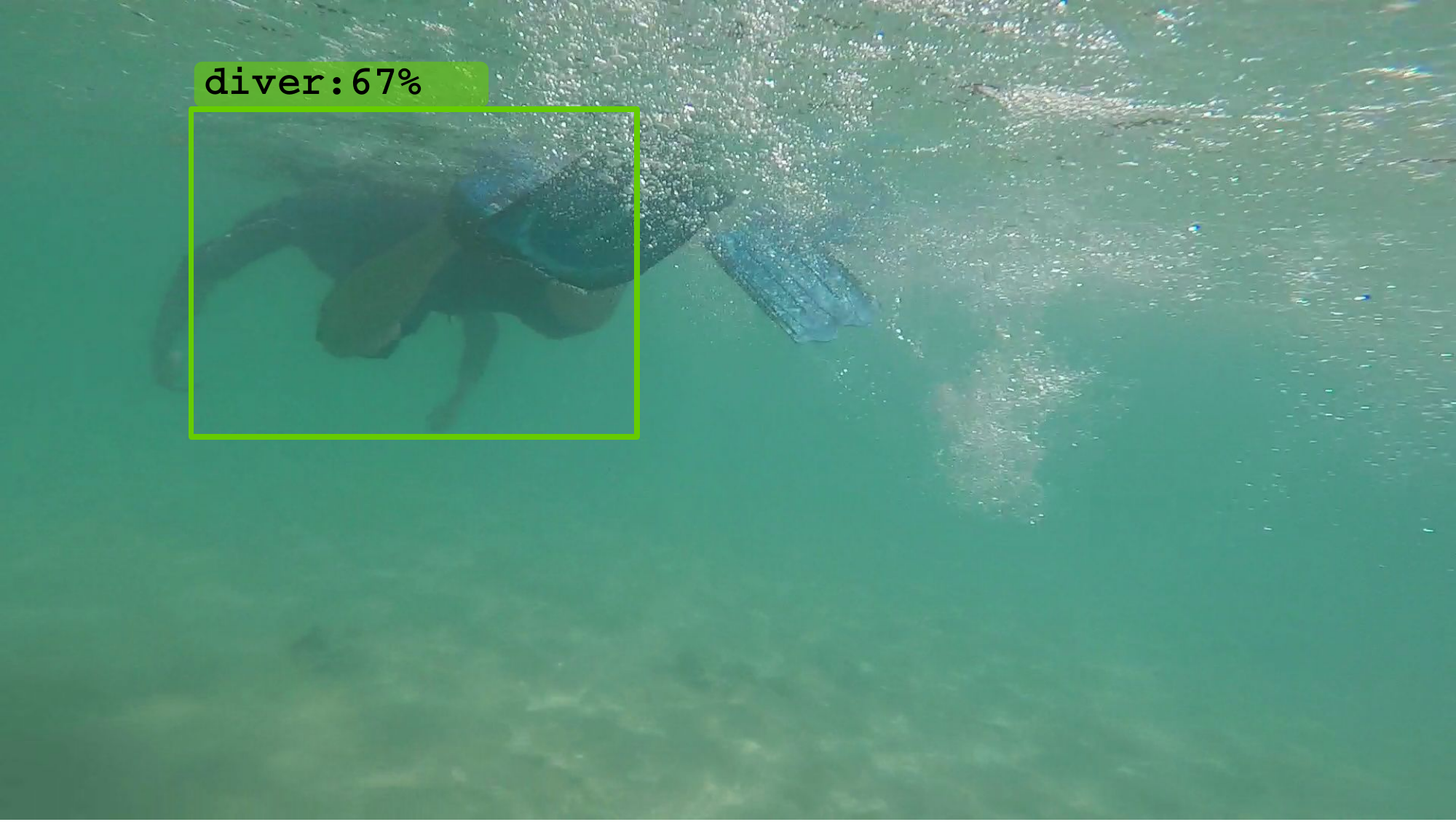}
        \caption{Air-bubbles produced from divers' flippers while swimming very close to the ocean surface}
        \end{subfigure}~
        \begin{subfigure}[t]{0.265\textwidth}
        \centering
        \includegraphics[width=\linewidth]{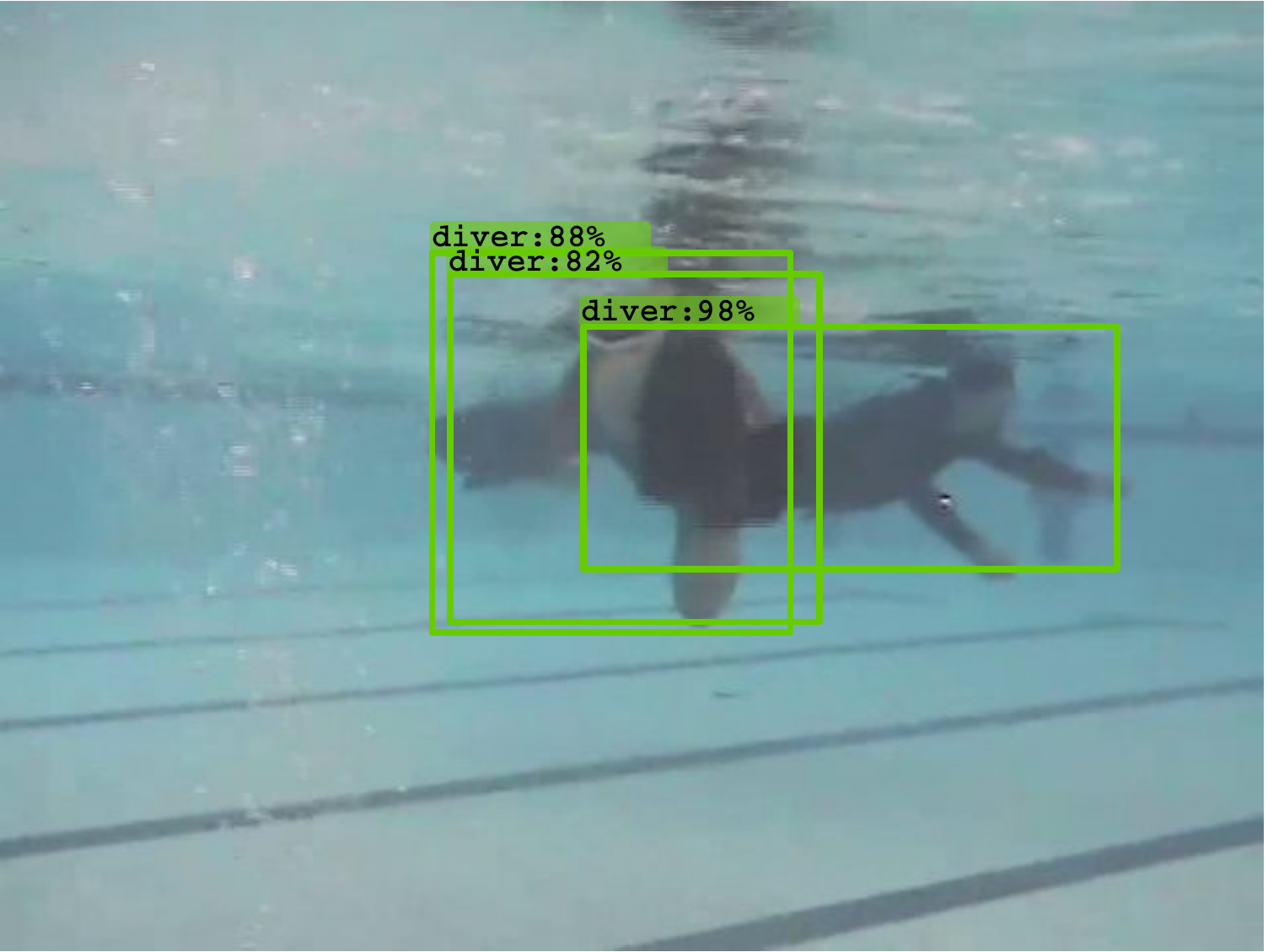}
        \caption{A diver is occluded by another}
        \end{subfigure}~
        \begin{subfigure}[t]{0.3\textwidth}
        \centering
        \includegraphics[width=\linewidth]{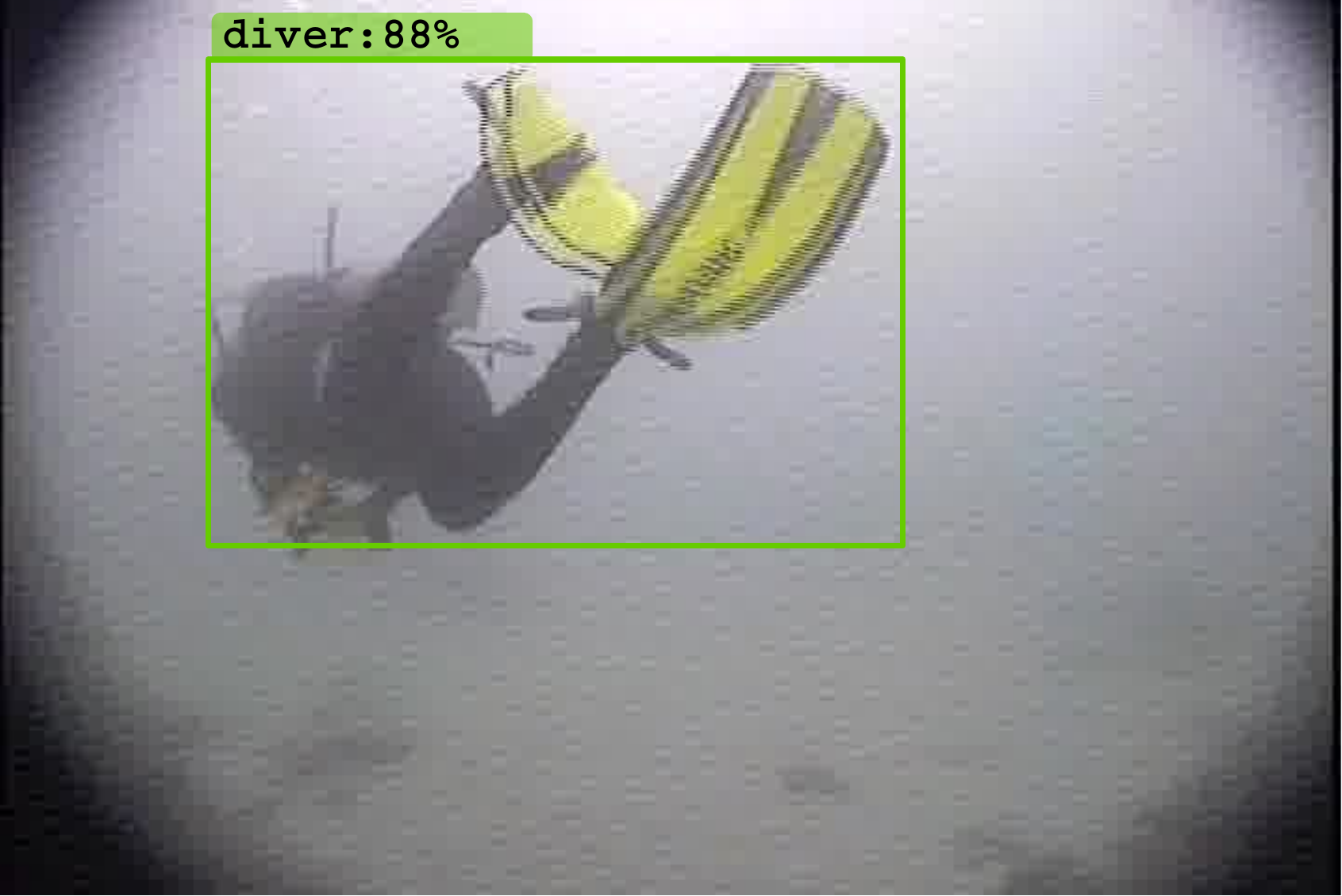}
        \caption{Color-distorted visuals due to poor lighting conditions}
        \end{subfigure}
     \caption{A few cases where the diver-detection performance is challenged by noise and occlusion. }
 \label{fig:bad}
\end{figure*}

\subsubsection{Feasibility and General Applicability}\label{field_res}
As mentioned, the diver-following module uses a monocular camera feed of the robot in order to detect a diver in the image-space and generate a bounding box. The visual servoing controller uses this bounding box and regulates robot motion commands in order to follow the diver. Therefore, correct detection of the diver is essential for overall success of the operation. We provided the detection performances of our proposed model over a variety of test scenarios in Table \ref{res_com} (few snapshots are illustrated in Figure \ref{fig:m1}). During the field experiments, we have found $6$-$7$ positive detections per second on an average, which is sufficient for successfully following a diver in real-time. In addition, the on-board memory overhead is low as the saved inference model is only about $60$MB in size. 

In addition, the proposed model is considerably robust to occlusion and noise, in addition to being invariant to divers' appearances and wearables. Nevertheless, the detection performances might be negatively affected by unfavorable visual conditions; we demonstrate few such cases in Figure \ref{fig:bad}. In Figure \ref{fig:bad}(a), 
the diver is only partially detected with low confidence ($67\%$). This is because the flippers' motion produces a flurry of air-bubbles (since he was swimming very close to the ocean surface), which occluded the robot's view. Suspended particles cause similar difficulties in diver-following scenarios. The visual servoing controller can recover from such inaccurate detections as long as the diver is partially visible. However, the continuous tracking might fail if the diver moves away from the robot's field of view before it can recover. In this experiment, $27$ consecutive inaccurate detections (\textit{i.e.}, confidence score less than $50\%$) caused enough drift in the robot's motion for it to lose sight of the person. On the other hand, occlusion also affects the detection performances as shown in Figure \ref{fig:bad}(b); here, the proposed model could not localize the two divers correctly due to occlusion.

\begin{figure}[!h]
 \centering
    \includegraphics[width=\linewidth]{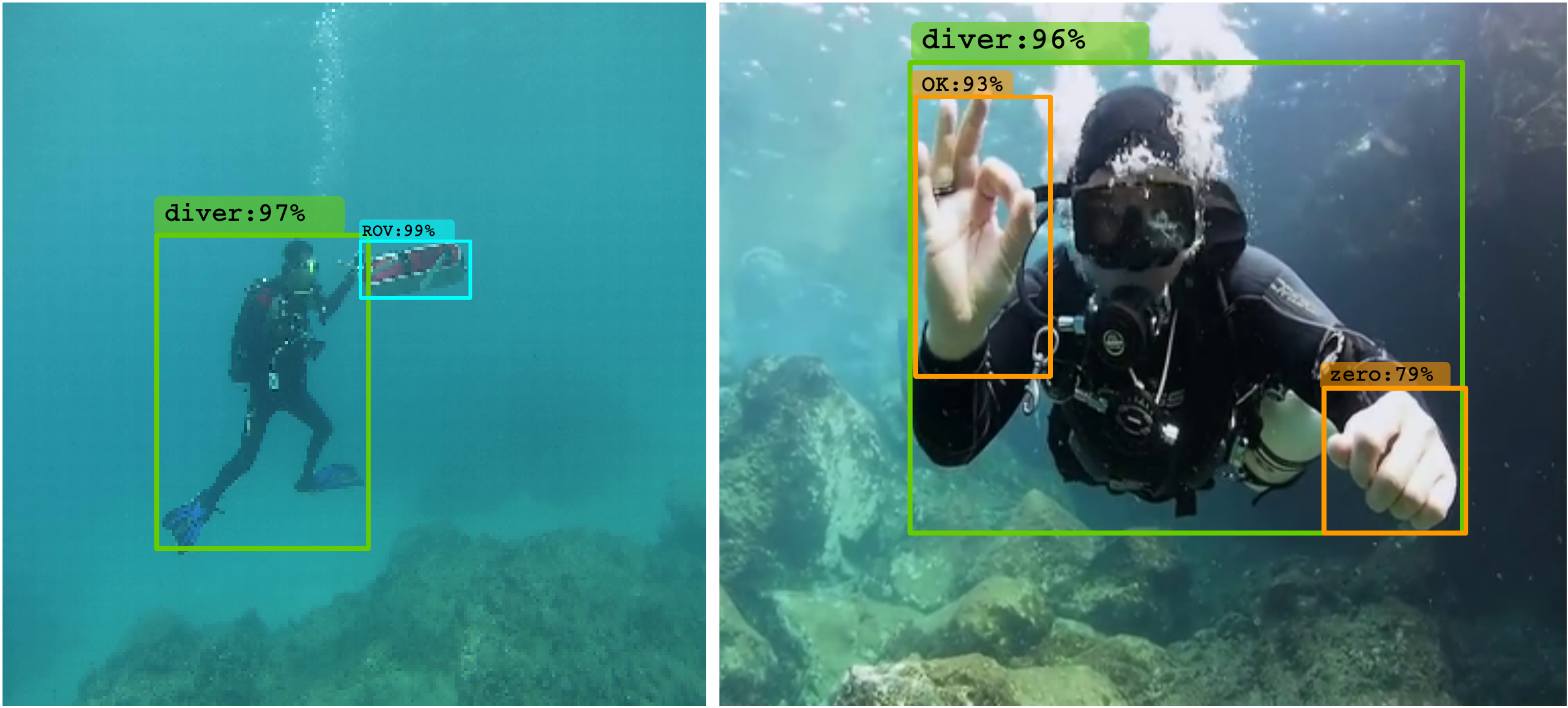}
 \caption{Detection of ROVs and hand gestures by the same diver-detector model. In this case, the SSD (MobileNet V2) model was re-trained on additional data and object categories for ROV and hand gestures (used for human-robot communication~\cite{islam2018dynamic}). }
 \label{fig:app}
\end{figure} 

Lastly, since our training datasets include a large collection of gray-scale and color distorted underwater images, the proposed models are considerably robust to noise and color distortions (Figure \ref{fig:bad}(c)). Nonetheless, state-of-the-art image enhancement techniques for underwater imagery  can be utilized to alleviate severe chromatic distortions. We refer interested readers to \cite{fabbri2018enhancing}, where we tried to address these issues for generic underwater applications.

We also performed experiments to explore the usabilities of the proposed diver detection models for other underwater applications. As demonstrated in Figure \ref{fig:app}, by simply re-training on additional data and object categories, the same models can be utilized in a wide range of underwater human-robot collaborative applications such as following a team of divers, robot convoying~\cite{shkurti2017underwater}, human-robot communication~\cite{islam2018dynamic}, etc. In particular, if the application do not pose real-time constraints, we can use models such as Faster R-CNN (Inception V2) for better detection performances.

%% file: srctex/con.tex
\section{CONCLUSION}
In this paper, we have tried to address the challenges involved in underwater visual perception for autonomous diver-following. 
At first, we investigated the performances and applicabilities of the state-of-the-art deep object detectors. We prepared and used a comprehensive dataset for training these models; then we fine-tuned each computational components in order to meet the real-time and on-board operating constraints. 
Subsequently, we designed a CNN-based diver detection model that establishes a delicate balance between robust detection performance and fast running time. Finally, we validated the tracking performances and general applicabilities of the proposed models through a number of field experiments in pools and oceans. 

In the future, we seek to improve the running time of the general object detection models on embedded devices. Additionally, we aim to investigate the use of human body-pose detection models to understand divers' motion, instructions, and activities.

\section*{ACKNOWLEDGMENT}
We gratefully acknowledge the support of the MnDirve initiative on this research. We are also thankful to the Bellairs Research Institute of Barbados for providing the facilities for our field experiments. In addition, we are grateful for the support of NVIDIA Corporation with the donation of the Titan Xp GPU for our research. We also acknowledge our colleagues, namely Cameron Fabbri, Marc Ho, Elliott Imhoff, Youya Xia, and Julian Lagman for their assistance in collecting and annotating the training data.